%% file: main.tex
\documentclass[preprint,12pt,numbers,compress]{elsarticle}

\usepackage{pifont}
\usepackage{amssymb}
\usepackage{amsmath}
\usepackage{subcaption}
\usepackage{booktabs}
\usepackage{float}
\usepackage{xcolor}
\usepackage{graphicx}
\usepackage{subcaption}
\usepackage{tikz}
\usepackage{pgfplots}
\usepackage{booktabs}
\usepackage{siunitx}

\pgfplotsset{compat=1.18} 

\definecolor{retinanet}{RGB}{64, 115, 158}    
\definecolor{ours}{RGB}{231, 76, 60}          
\definecolor{avgcolor}{RGB}{46, 204, 113}     
\definecolor{highlight}{RGB}{255, 127, 0}     

\journal{Expert Systems with Applications}

\begin{document}

\begin{frontmatter}



\title{RMK RetinaNet: Rotated Multi-Kernel RetinaNet for Robust Oriented Object Detection in Remote Sensing Imagery}

\author[a]{Huiran Sun\corref{cor1}}
\affiliation[a]{organization={School of Computer Science and Engineering, Changchun University of Technology},
	addressline={2055 Yan'an Street, Chaoyang District}, 
	city={Changchun},
	postcode={130012}, 
	state={Jilin Province},
	country={China}}

\cortext[cor1]{Corresponding author:Huiran Sun,\textbf{Email:} 
}

\begin{abstract}
Rotated object detection in remote sensing imagery is hindered by three major bottlenecks: non-adaptive receptive field utilization, inadequate long-range multi-scale feature fusion, and discontinuities in angle regression. To address these issues, we propose Rotated Multi-Kernel RetinaNet (RMK RetinaNet). First, we design a Multi-Scale Kernel (MSK) Block to strengthen adaptive multi-scale feature extraction. Second, we incorporate a Multi-Directional Contextual Anchor Attention (MDCAA) mechanism into the feature pyramid to enhance contextual modeling across scales and orientations. Third, we introduce a Bottom-up Path to preserve fine-grained spatial details that are often degraded during downsampling. Finally, we develop an Euler Angle Encoding Module (EAEM) to enable continuous and stable angle regression. Extensive experiments on DOTA-v1.0, HRSC2016, and UCAS-AOD show that RMK RetinaNet achieves performance comparable to state-of-the-art rotated object detectors while improving robustness in multi-scale and multi-orientation scenarios.
\end{abstract}




\end{frontmatter}



\section{Introduction}

\begin{figure}[!t]
	\captionsetup{labelsep=period}
	\raggedleft 
	\begin{minipage}[t]{1\textwidth} 
		\centering
		\includegraphics[width=\linewidth]{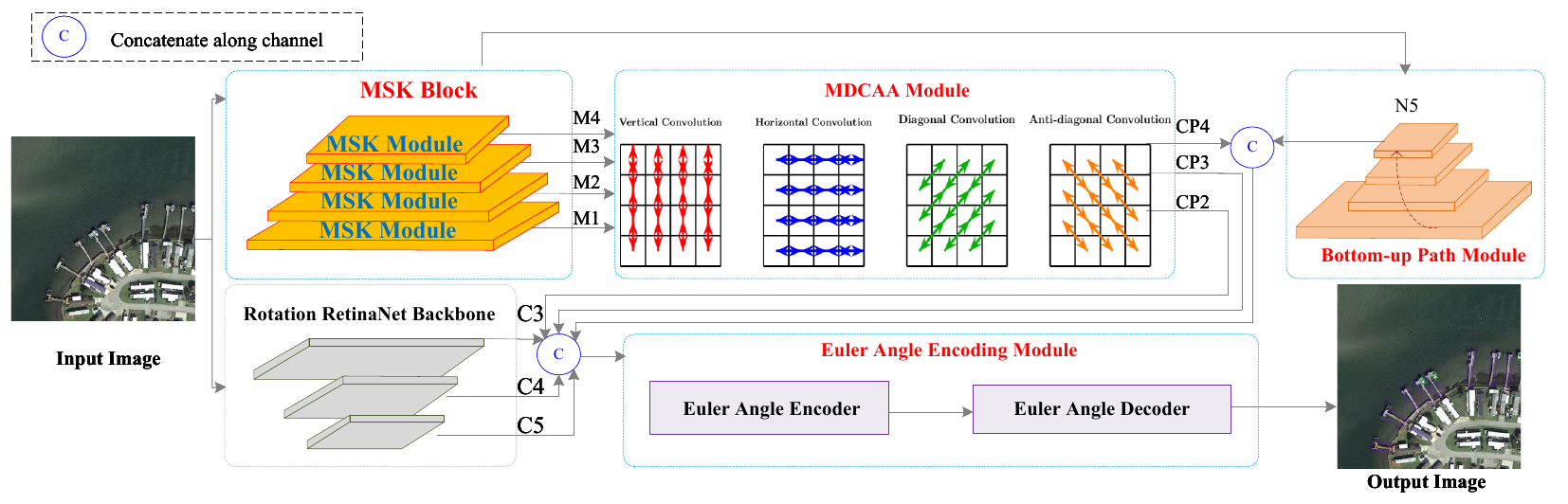}
		\caption{\textbf{Overview of RMK RetinaNet.} The \textbf{MSK Block} consists of four layers, each constructed by the \textbf{MSK Module} \textcolor{red}{(\S\ref{subsec:MSK Module})}.
			Only the bottommost MSK module does not perform downsampling; the remaining three modules all do.The outputs of the MSK Module are labeled as $M_1$, $M_2$, $M_3$, and $M_4$, which serve as inputs to both the \textbf{MDCAA Module}\textcolor{red}{(\S\ref{subsec:MDCAA Module})} and the \textbf{Bottom-up Path Module} \textcolor{red}{(\S\ref{subsec:Bottom-up Path Module})}. $M_1$, $M_2$, $M_3$, and $M_4$ are fed into the MDCAA Module to generate $CP_2$, $CP_3$, and $CP_4$. Similarly, when fed into the Bottom-up Path, only the top-level output $N5$ is retained. Feature maps $CP_4$, $N5$, $CP_3$, and $CP_2$ are then concatenated with the $C_5$, $C_4$, and $C_3$ feature maps from Rotation RetinaNet at corresponding scales. Finally, the combined features are processed by the \textbf{Euler Angle Encoding Module} \textcolor{red}{(\S\ref{subsec:Euler Angle Encoding Module})} to produce the final detection results.}
		\label{fig:net-overview}
	\end{minipage}
	\vspace{-10pt} 
\end{figure}

Object detection in remote sensing imagery~\cite{gao2026explicitly,rajendran2025adaptive,zhao2025differential,dong2024elnet,sagar2024msa,xiao2023fdlr,chen2023infofpn} has attracted growing attention in recent years. The goal is to identify target instances and predict their categories and spatial locations. Different from generic detection that typically outputs horizontal bounding boxes, remote sensing detection often requires \emph{rotated} bounding boxes to better align with arbitrarily oriented objects. Accordingly, a series of rotated detectors have been developed, such as Rotated CenterNet~\cite{cheng2023oriented} and Rotation RetinaNet~\cite{zan2019automatic}. Although many previous works have focused on developing various oriented bounding box (OBB) detectors~\cite{han2021align,li2022oriented,xie2021oriented,xu2020gliding} and improving the angular prediction accuracy of OBBs~\cite{yang2021dense,yang2021rethinking,yang2022kfiou}, effective feature extraction tailored to the unique characteristics of remote sensing imagery remains comparatively less explored.

Remote sensing images are captured from a bird's-eye view by platforms such as satellites and unmanned aerial vehicles (UAVs). With the rapid progress of aerospace and sensing technologies, the spatial resolution and visual quality of remote sensing imagery have continuously improved, enabling a wide range of applications including environmental monitoring, urban planning, maritime surveillance, and military reconnaissance. However, remote sensing scenes pose several intrinsic challenges beyond those in natural images. Targets (e.g., ships, vehicles, aircraft, and buildings) exhibit arbitrary orientations, dense distributions, cluttered backgrounds, and, importantly, extreme scale variations---from large objects such as playgrounds to small instances such as vehicles. Recognizing large-scale targets often relies on contextual cues and global appearance, whereas detecting small objects depends more on fine-grained spatial details. Although multi-scale feature fusion strategies (e.g., FPN-style designs) are widely adopted to mitigate scale variation, they still share several limitations in remote sensing scenarios.

First, receptive field designs are typically \emph{non-adaptive}: a fixed receptive field is applied to objects with substantially different sizes, leading to insufficient contextual coverage for large targets and suboptimal modeling of their spatial structures. Second, existing feature pyramids usually perform feature addition or concatenation only between adjacent levels, lacking structured interaction across \emph{distant} scales. This limits effective collaboration between deep semantic features and shallow high-resolution details, thereby constraining multi-scale fusion efficiency. Third, many rotated detectors remain affected by periodicity and ambiguity in angle regression. Near boundary regions of the angle parameterization, the loss may become discontinuous, causing gradient oscillation and optimization instability, which further degrades localization robustness, especially in dense scenes.

Recent advances in large-kernel modeling~\cite{ding2022scaling,guo2023visual,liu2022more,liu2022convnet} have provided a practical direction to enlarge receptive fields. LSKNet~\cite{li2023large} expands receptive fields via large-kernel and dilated convolutions to capture richer context, yet large kernels may introduce background noise that harms small-object detection, and dilated convolutions can miss local details due to sparse sampling. As an evolution, PKINet~\cite{cai2024poly} employs parallel multi-scale depthwise separable convolutions to extract dense multi-receptive-field features, and introduces contextual anchor attention for long-range dependency modeling. Despite its effectiveness, PKINet still adopts element-wise addition for feature fusion, which is often overly linear and may reduce discriminability, blur details, and amplify noise---issues that are particularly severe in remote sensing imagery with complex backgrounds and strong scale variation. Nevertheless, PKINet provides a strong feature extraction backbone that jointly offers multi-scale semantic cues and spatial geometric fidelity, which is valuable for subsequent rotated detection.
On the other hand, Oriented R-CNN alleviates the periodic discontinuity in angle regression by transforming the rotation angle into directed distances from vertices of the enclosing horizontal box to the edges of the rotated box. This converts angle regression with periodic singularities into smooth distance regression in Euclidean space, improving training stability. However, this strategy mainly avoids practical optimization issues and does not fundamentally address the discontinuity and ambiguity inherent to angle parameterization, indicating that further improvements are still possible in both feature extraction and geometric modeling for rotated detection.
To this end, we propose Rotated Multi-Kernel RetinaNet (RMK RetinaNet) for rotated object detection in remote sensing imagery. RMK RetinaNet is built upon four key components:

\begin{itemize}
\item  To overcome fixed receptive fields, we introduce a multi-kernel parallel perception strategy. Specifically, standard 2D convolutions are decomposed into multi-scale, orthogonal 1D convolution sequences, enabling directional and shape-aware receptive field adaptation with reduced parameter redundancy. This design preserves geometric details of slender and rotated targets while mitigating background clutter.
\item  To strengthen orientation perception under cluttered backgrounds, we design MDCAA Module that leverages global semantics as anchors and integrates multi-directional (horizontal, vertical, and diagonal) strip convolutions to model long-range dependencies across the image. By injecting explicit directional context, MDCAA Module dynamically re-weights target-relevant features, suppressing background noise and improving discrimination for elongated and rotated objects.
\item We augment the classic FPN with a Bottom-up Path to better preserve fine-grained spatial details that are weakened by repeated downsampling. Low-level positional cues are injected into the fusion process, improving localization accuracy, particularly for small targets.
\item  To address boundary discontinuity in angle regression, we propose a continuous and differentiable reversible Euler Angle Encoding Module that maps angles to 2D unit-circle vectors. This naturally eliminates periodic jumps and reduces orientation ambiguity, improving convergence stability and robustness in dense and complex scenes.
\end{itemize}

\section{Related Work}

\subsection{Oriented Object Detection in Remote Sensing Imagery}
In oriented object detection, objects are commonly represented by the minimum enclosing rotated rectangle, i.e., the oriented bounding box (OBB), parameterized as $(x,y,w,h,\theta)$, where $(x,y)$ denotes the box center, $(w,h)$ are the width and height, and $\theta$ is the rotation angle. Most rotated detectors are extended from horizontal detection frameworks to better handle arbitrary orientations and dense layouts in remote sensing scenes. Representative two-stage approaches include ROI-Transformer~\cite{ding2019learning}, SCRDet~\cite{yang2019scrdet}, and ReDet~\cite{han2021redet}, while popular single-stage pipelines cover DRN~\cite{zheng2018drn}, R3Det~\cite{ma2022recognition}, and S2A-Net~\cite{li2023improved}. Despite notable progress, their performance is still constrained by challenges intrinsic to remote sensing imagery, including extreme scale variation, cluttered backgrounds, and the instability of angle regression near periodic boundaries.
Beyond detection, recent studies in controllable generation and editing emphasize that robust geometric modeling and stable optimization often depend on well-behaved representations and lightweight plug-in modules, which can significantly improve controllability and robustness without redesigning the entire framework~\cite{shen2025imagedit,shen2025imagharmony,shen2025imaggarment}. Similar design philosophy has also been validated in pose-/motion-conditioned generation tasks, where continuous conditioning and stable parameterization are essential for long-horizon consistency and reliable convergence~\cite{shen2024imagpose}.

\subsection{Angle Parameterization and Boundary Discontinuity}
The \emph{boundary discontinuity problem} is a long-standing bottleneck in oriented object detection. In horizontal detection, IoU-based regression losses are widely adopted, yet early oriented detectors often avoided rotated IoU due to its computational complexity and non-differentiability, and instead regressed $(x,y,w,h,\theta)$ independently using Smooth L1 loss. However, due to the periodic nature of $\theta$, the loss can exhibit abrupt jumps when angles cross boundary regions, inducing gradient oscillations and training instability. CSL~\cite{yang2020arbitrary} explicitly formalized this phenomenon and identified angle periodicity as the key cause.
More broadly, continuous and reversible encoding has been repeatedly shown to improve optimization stability and reduce ambiguity in geometry-related prediction tasks, especially under complex supervision signals and long-range dependencies~\cite{shen2024advancing,shen2025imagdressing}. Motivated by this line of work, we introduce EAEM as a continuous angle encoding strategy to mitigate periodic discontinuity and enhance orientation regression stability in dense and complex remote sensing scenes.

\section{Method}
\begin{figure}[!t]
	\captionsetup{labelsep=period}
	\centering
	\begin{minipage}[t]{1.0\textwidth}
		\centering
		\includegraphics[width=\linewidth]{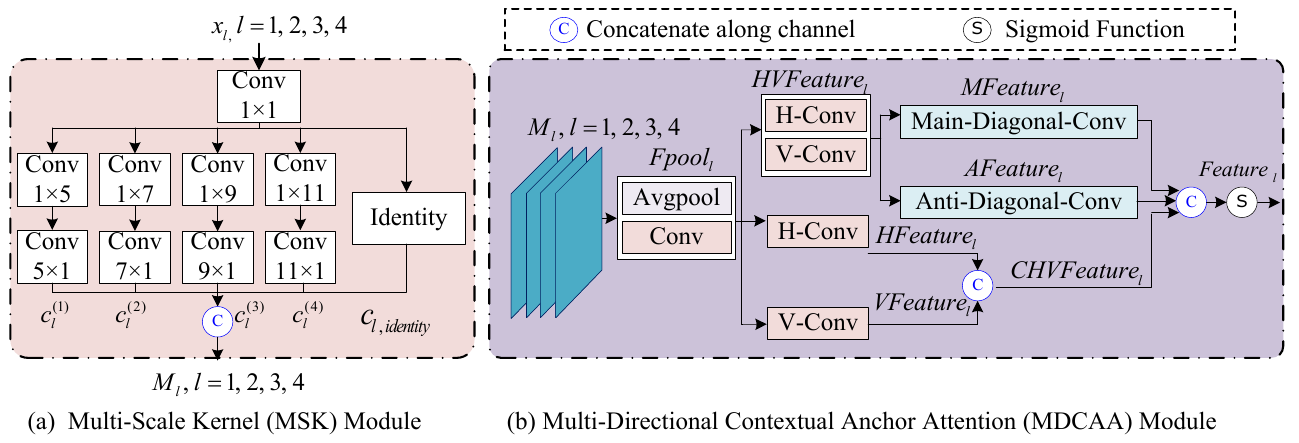}
		\caption{
			\textbf{(a)} The MSK Block consists of four MSK Modules. Here, we use \( l \) to denote the \( l \)-th \textbf{MSK Module} \textcolor{red}{(\S\ref{subsec:MSK Module})}, and the output of each MSK Module is denoted as $M_l$.\\
			\textbf{(b)}In the \textbf{MDCAA Module}\textcolor{red}{(\S\ref{subsec:MDCAA Module})}, horizontal convolution is denoted as $H$-$Conv$, vertical convolution as $V$-$Conv$, convolution along the main diagonal direction as $Main$-$Diagonal$-$Conv$, and convolution along the anti-diagonal direction as $Anti$-$Diagonal$-$Conv$.
		}
		\label{fig:MSK Module and MDCAA Module}
	\end{minipage}
	\vspace{-10pt}
\end{figure}
\subsection{Rotated Multi-Kernel RetinaNet}
\label{subsec:RMK RetinaNet}
As shown in Figure 1, our RMK RetinaNet is a backbone network that integrates the MSK Block with the base network RetinaNet. The output of the MSK Block is divided into two branches. In the first branch, we design the MSK Block, which consists of four MSK modules (details in Section 3.2). Its output serves as the input to the MDCAA Module (details in Section 3.3). At this stage, the MDCAA mechanism is designed to effectively capture the directional sensitivity present in the features, thereby improving feature quality. The output of the second branch of the MSK Block is fed into the Bottom-up Path Module(details in Section 3.4) to improve the localization capability of low-level features. Finally, the outputs from these two branches are concatenated and fused with the feature maps $C5$, $C4$, and $C3$ from the baseline RetinaNet path, and then fed into the Euler Angle Encoding Module (details in Section 3.5) to jointly generate the final output features for the current stage. Through this dual-path collaboration and information complementarity design, RMK RetinaNet effectively balances semantic richness and localization accuracy, providing more robust feature representations for object detection tasks.
\subsection{MSK Module}
\label{subsec:MSK Module}
The MSK Block consists of four MSK modules, and the structure of each MSK Module is shown in Figure 2(a). In the multi-scale kernel Module proposed in this study, an Inception-like structure is innovatively introduced. Four independent standard convolution branches are deployed in parallel, with convolution kernel sizes of  $m \in \{5, 7, 9, 11\}$ respectively, enabling the network to simultaneously capture cross-scale feature representations ranging from local texture features to global contextual information. The l-th MSK Module can be mathematically expressed as follows:
\begin{equation}
	 \mathbf{X}_{l,identity}= \mathrm{Conv}_{3 \times 3}\left(\mathrm{Conv}_{1 \times 1}\left(\mathbf{X}_{l}\right)\right), \quad l = 1, \dots, 4,
\end{equation}
\begin{equation}
	\begin{split}
		\mathbf{X}_{l}^{(k)} = \mathrm{Conv}_{m \times 1}\left( \mathrm{Conv}_{1 \times m}\left( \mathrm{Conv}_{1 \times 1}\left( \mathbf{X}_{l} \right) \right) \right), \\
		\text{where } l = 1, \dots, 4; \; k = 1, \dots, 4; \; m \in \{5, 7, 9, 11\},
	\end{split}
\end{equation}
Here, $X_{l, identity}$ $\in$ $\mathbb{R}^{C \times H \times W}$; in formula (1), the number of channels of the feature map $X_{l}$ is first adjusted using a $1 \times 1$ convolution, followed by local feature extraction with a $3 \times 3$ convolution. In formula (2), $X_{l}^{(k)}$ $\in$ $\mathbb{R}^{C \times H \times W}$, the feature map $X_{l}$, after being similarly adjusted in channel number using a $1 \times 1$ convolution, then two strip convolutions of size $1 \times m$ and $m \times 1$ are connected in series to replace the standard two-dimensional $m \times m$ convolution. Considering the design of the PKI Module in PKINet, although it can extract features with different receptive field scales, the resulting parameter and computational costs are also significant enough to warrant attention. Therefore, we decided to adopt the spatially separable convolution technique, replacing the standard two-dimensional $m \times m$ convolution with two strip convolutions of size $1 \times m$ and $m \times 1$.For an $m \times m$ convolution kernel, the number of parameters is $C_{in} \times C_{out} \times m \times m$. However, if it is split into two kernels of size $1 \times m$ and $m \times 1$ while maintaining the same receptive field, the total number of parameters becomes $C_{in} \times C_{mid} \times 1 \times m + C_{mid} \times C_{out} \times m \times 1$. If we set $C_{mid} = C_{out} = C_{in}= C$, a simple calculation shows that the parameter count for the standard two-dimensional convolution is $C^{2}m^{2}$, while that for the spatially decomposed convolution is $2C^{2}m$. Comparing the latter to the former gives a parameter ratio of $2/m$. This ratio formula of $2/m$ is derived under the assumption that the input, output, and intermediate channel numbers are all equal, reflecting the theoretical advantage of spatially separable convolution in parameter efficiency. Then, using formula (3),
\begin{equation}
	\mathbf{M}_{l} =\mathrm{Concat} \left( \mathbf{X}_{l}^{(k)}, \mathbf{X}_{l,identity} \right), \text{where } l = 1, \dots, 4; \; k = 1, \dots, 4. \tag{3}
\end{equation}
$X_{l, identity}$ and $X_{l}$ are concatenated along the first dimension, capturing the interrelationships between different channels.

Table 1 presents a comparison between our proposed MSK Module and the baseline PKI Module in terms of model complexity. Under the condition of an input image size of $ 1024 \times 1024 $, the computational cost (FLOPs) of the two models remains consistent, while the number of parameters in the MSK Module experiences a slight reduction of 0.054M (from 33.612M to 33.558M). It is important to emphasize that the parameter count (Params) is an inherent property of the model, and this comparative result is independent of the number of input images. The reduction in parameters shown in the table 1 signifies a permanent decrease in the model's intrinsic storage overhead. Although the magnitude of reduction per image may appear modest, its benefits become significantly amplified when extended to the training and deployment scenarios involving the entire dataset:
\begin{itemize}
	\item During the training phase, the reduction in parameters directly decreases GPU memory consumption, as each parameter requires storage for its value, gradient, and optimizer states (e.g., momentum and variance in Adam). This facilitates the use of larger batch sizes under the same hardware constraints or enables model deployment on edge devices with more limited memory.
    \item During the inference phase, the smaller model size facilitates faster model loading and transmission, which holds practical significance for large-scale service deployment or mobile applications.
\end{itemize}
Therefore, the implementation of the MSK Module introduces a parameter-efficient model optimization. Its lightweight advantage becomes more pronounced in practical applications involving massive data processing.
\begin{table}[t]
	\centering
	\small  %
	\setlength{\tabcolsep}{0.34pt} 
	\label{tab:results}
	\input{./Tables1/table1.tex} 
\end{table}
In addition to its parameter-light design, the MSK Module incorporates corresponding improvements in its feature fusion strategy. As illustrated in Figure 2(a), the feature maps output by each branch are fused via channel-wise concatenation, as opposed to the traditional element-wise addition of feature maps.
This design choice is primarily motivated by the observation that in remote sensing image analysis, effective detection of objects at different scales relies on their distinct feature representations. Element-wise feature addition smooths feature responses through numerical summation, which may weaken the discriminative features extracted by each branch. In contrast, channel concatenation fully preserves the original feature distribution extracted by each scale branch by merging them in parallel along the channel dimension, thereby forming a multi-scale feature descriptor with stronger discriminative power. Moreover, from the perspective of representation learning theory, channel concatenation effectively expands the representational capacity of the feature space by increasing the channel dimension of the feature maps. This dimensional expansion provides richer feature bases for subsequent network layers, enabling the network to construct more discriminative composite features through hierarchical learning, which is of significant importance for handling the complex background interference commonly encountered in remote sensing images.
\subsection{MDCAA Module}
\label{subsec:MDCAA Module}
In remote sensing object detection, targets often exhibit strong directional characteristics (such as ships, airplanes, etc.), and their discriminative contextual information may be distributed in arbitrary orientations. Traditional attention mechanisms (e.g., SE, CBAM) primarily focus on channel-wise or axial attention, making it difficult to effectively capture such complex spatial contexts. Inspired by Contextual Anchor Attention (CAA), this paper proposes an MDCAA mechanism, the structure of which is illustrated in Figure 2(b). The MDCAA Module is designed to overcome the directional limitations of traditional convolutional kernels, simulating the ability of the human visual system to capture contextual information from vertical, horizontal, and oblique diagonal directions.

We first apply average pooling to each input feature map, followed by a pointwise convolution to extract local region features, yielding a set of pooled feature representations. Subsequently, we apply two deep separable strip convolutions to approximate a standard large-kernel deep separable convolution. Specifically, we first apply a vertical strip depthwise separable convolution followed by a horizontal strip depthwise separable convolution to the pooled features, where the kernel size in the strip direction is denoted by a parameter $m$. This operation yields a feature map that captures contextual information along both orientations, denoted as ${HVFeature}_{l}$. Furthermore, using separate branches, we independently apply horizontal strip depthwise separable convolution and vertical strip depthwise separable convolution to the pooled features. This produces distinct feature maps: ${HFeature}_{l}$ which emphasizes horizontal context, and ${VFeature}_{l}$ which emphasizes vertical context. Moreover, the feature map ${HFeature}_{l}$ obtained from the horizontal strip convolution and the feature map ${VFeature}_{l}$ obtained from the vertical strip convolution are concatenated along the channel dimension to generate a combined feature map, denoted as ${CHVFeature}_{l}$, which integrates contextual information from both orientations.

To enhance the model's perception capability for directional features, we propose a diagonal convolution method based on spatial rotation transformation. The feature maps obtained from Equation (4) and Equation (5) are rotated 90 degrees counterclockwise or clockwise, converting diagonal spatial patterns into axial patterns. Standard convolution kernels are then utilized for efficient feature extraction. Finally, the feature map is restored to its original orientation through inverse rotation to maintain the consistency of spatial relationships:
\setcounter{equation}{3}
\begin{align}
	\mathbf{MFeature}_{l} & = \mathrm{Rot}_{-90^{\circ}}(\mathbf{HVFeature}_{l}), \\
	\mathbf{AFeature}_{l} & = \mathrm{Rot}_{+90^{\circ}}(\mathbf{HVFeature}_{l}),
\end{align}
Here, ${Rot}_{-90^\circ}$ denotes a clockwise rotation by 90 degrees, and ${Rot}_{+90^\circ}$ denotes a counterclockwise rotation by 90 degrees. This operation enables the model to directly capture long‑range contextual dependencies along the potential diagonal orientation of the target.
Finally, the feature maps ${MFeature}_{l}$, ${AFeature}_{l}$, and 
${CHVFeature}_{l}$ are first concatenated along the channel dimension. A $1 \times 1$ convolution is then applied to ensure the fusion of the concatenated features, followed by a Sigmoid function to generate the MDCAA attention weights, which are constrained between 0 and 1.
\subsection{Bottom-up Path Module}
\label{subsec:Bottom-up Path Module}
In feature pyramid networks, the precise localization information abundant in low-level feature maps continuously attenuates and becomes blurred during propagation to higher layers due to repeated downsampling. To address this critical deficiency, this paper innovatively introduces a Bottom-up Path Module. This Module takes the output feature pyramid $\{\mathbf{M}_l\}_{l=1}^4$ from the MSK Module as input. Its core idea is to start from the \(\mathbf{M}_1\) layer with the highest spatial resolution and construct a path that transmits high-resolution details from lower to higher layers through a series of convolution downsampling operations with a stride of 2. The computational process of this path can be expressed by Equation (12) as follows:
\begin{equation}
	\begin{split}
		N_{i} = \mathrm{Conv}_{3 \times 3} \bigl( 
		& \mathbf{M}_{l+1} + \\
		& \mathrm{DownSample}(\mathbf{M}_{l}) \bigr), \\
		& l=1,2,3, \quad i = 2, \dots, 5,
	\end{split}
\end{equation}

Among them, ${DownSample}$ represents a downsampling convolution with a stride of 2. Its function is to halve the size of the feature map at layer $l$ so that it aligns with the size of the feature map at layer $l+1$. ${Conv}_{3 \times 3}$ is a $3\times3$ convolution applied to the corresponding feature map at each layer, which refines the fused features. The "\(+\)" operation achieves element-wise fusion of the semantic information from the feature pyramid with the positional information from the Bottom-up Path.

Finally, the Bottom-up Path Module takes $N_5$ as an enhanced high-level feature rich in precise location information, which is used for subsequent path feature aggregation, laying a solid foundation for generating high-quality detection results.
\subsection{Euler Angle Encoding Module}
\label{subsec:Euler Angle Encoding Module}
We design an Euler Angle Encoding Module that adjusts the rotation angle by constructing a continuous, differentiable, and invertible Euler angle encoding function. Its specific definition is as follows:

\textbf{Definition 1:} Given Euler's formula \( z = \cos(\omega\theta) + j\sin(\omega\theta) \), let \( x = \cos(\omega\theta) \), \( y = \sin(\omega\theta) \),and impose the constraint: on the unit circle \( x^2 + y^2 = 1 \), and \( \omega\theta \in [0,2\pi) \).

Here, \( z \) is the encoded value, \( j \) denotes the imaginary unit, and \( \omega \in \mathbb{R}^+ \) is the angular frequency. Decoding can only obtain a unique angle within one period on the complex plane, with the range of \( \omega\theta \) being \( [0,\omega\pi) \subseteq [0,2\pi) \).Therefore, it is necessary to satisfy \( \omega \leq 2 \) .

\textbf{Definition 2:}The inverse mapping of Euler's formula is given by 
\[
\theta = \frac{1}{\omega} \, \mathrm{Arg}_{[0,2\pi)}(z),
\]
where the function \(\mathrm{Arg}_{[0,2\pi)}(z)\) is defined as follows:
\[
\mathrm{Arg}_{[0,2\pi)}(z) = 
\begin{cases} 
	\mathrm{arctan}\left(\dfrac{y}{x}\right), & x > 0,\; y \geq 0, \\[10pt]
	\mathrm{arctan}\left(\dfrac{y}{x}\right) + 2\pi, & x > 0,\; y < 0, \\[10pt]
	\mathrm{arctan}\left(\dfrac{y}{x}\right) + \pi, & x < 0, \\[10pt]
	\dfrac{\pi}{2}, & x = 0,\; y > 0, \\[10pt]
	\dfrac{3\pi}{2}, & x = 0,\; y < 0, \\[10pt]
	\text{undefined}, & x = 0,\; y = 0.
\end{cases}
\]
In this piecewise function, we discuss the following six cases:

(1) When \( x > 0 \), the point lies in the first or fourth quadrant. The angle is in \( \left(-\pi/2, \pi/2\right) \). For \( y \geq 0 \), we directly take \(\arctan\left(y/x\right) \in \left[0,\pi/2\right)\);

(2) For \( x > 0 \), \( y < 0 \), \(\arctan\left(y/x\right) \in \left(-\pi/2, 0\right)\), and we add \( 2\pi \) to obtain \(\left[3\pi/2, 2\pi\right)\);

(3) When \( x < 0 \), regardless of the value of \( y \), the point lies in the second or third quadrant. \(\arctan\left(y/x\right) \in \left(-\pi/2, \pi/2\right)\), and adding \(\pi\) yields \(\left(\pi/2, 3\pi/2\right)\), which is the correct interval.

(4) When \( x = 0 \), \( y > 0 \), special handling is required: the positive y-axis corresponds to \(\pi/2\);

(5) When \( x = 0 \), \( y < 0 \), special handling is required: the negative y-axis corresponds to \(3\pi/2\);

(6) Since when \( x = 0, y = 0 \), we have imposed the constraint in Definition 1, \( x^2 + y^2 = 1 \), this case cannot occur.

Since the introduced Euler formula encodes angles as continuous coordinates on the unit circle in the complex plane, it enables smooth transitions for rotated bounding box angle regression at periodic boundaries, eliminating the numerical jumps and loss discontinuities inherent in traditional angle parameterization. This results in more stable and efficient gradient descent optimization. Furthermore, its invertible encoding-decoding structure ensures that the target orientation angle can be uniquely and accurately recovered from the two-dimensional vector output by the network, avoiding ambiguity in direction prediction. Ultimately, by transforming discrete angular jumps into continuous sliding along the unit circle and converting periodic discontinuities into closed continuous cycles, this complex exponential-based representation significantly enhances the angular modeling capability of rotated object detection in remote sensing imagery. This allows the detection outcomes to maintain high accuracy while achieving improved numerical stability and robustness.
\section{Experiment}
In this section, we will detail the experimental setup and results analysis, demonstrate the performance and efficiency of the RMK RetinaNet model, and compare it with multiple existing methods on challenging computer vision object detection tasks.
\subsection{Dataset}
The DOTAv1.0 dataset, jointly released by the Computer Vision and Remote Sensing Laboratory of Wuhan University and several remote sensing institutions, is a large-scale dataset specifically designed for oriented bounding box detection in aerial imagery. The dataset is collected from multiple regions worldwide (Asia, North America, and Europe), covering various geographical environments such as urban areas, rural landscapes, and coastal zones. It includes 15 different object categories: plane, baseball diamond, bridge, ground track field, small vehicle, large vehicle, ship, tennis court, basketball court, storage tank, soccer ball field, roundabout, harbor, swimming pool, and helicopter. DOTAv1.0 comprises a total of 2,806 aerial images with 188,282 oriented bounding box annotations across the 15 object categories. The training set, validation set, and test set contain 1,411, 458, and 937 images respectively, making it particularly suitable for the following tasks: oriented object detection of ships and planes, as well as small object detection for targets occupying only a few tens of pixels, such as vehicles and helicopters. Thanks to its extensive category coverage, diverse image backgrounds, and rich variation in object orientations, DOTAv1.0 has become a key benchmark dataset in the field of remote sensing image analysis. We crop the training images into $1024\times 1024$ pixel tiles with a 256-pixel overlap between adjacent tiles. During inference, the test set images are divided into 1024×1024 pixel tiles with a 200-pixel overlap between adjacent tiles to mitigate boundary artifacts caused by tiling operations.

The HRSC2016 dataset provides high-resolution ship imagery covering various operational scenarios, including maritime navigation and nearshore operations. The dataset has a resolution range between 0.4 meters and 2 meters, with ship image sizes varying from 300 to 1500 pixels, and most images larger than 1000×600 pixels. We crop the training images into 640×640 pixel patches with a 160-pixel overlap between adjacent patches. During testing, the test set images are cropped into $1024\times 1024$ pixel patches with a 200-pixel overlap between adjacent patches.

The UCAS-AOD dataset contains two categories, cars and airplanes, with a total of 1,510 aerial images at a resolution of approximately 659×1280 pixels, comprising 14,596 annotated instances. We randomly selected 1,110 samples for training and 400 samples for testing. Ultimately, the same data processing strategy as used for HRSC2016 was adopted.

\subsection{Experiment Details}
All methods in this paper are implemented using PyTorch and trained on NVIDIA RTX 3090 GPUs. We adopt the proposed RMK RetinaNet as the network architecture, employing a pre-trained ResNet-50 as the backbone network and utilizing the EAEM method for angle encoding and decoding of rotated objects. To achieve high detection accuracy while ensuring a fair comparison, the batch size is set to 12 and the initial learning rate is set to $1.25 \times 10^{-4}$ when training on the DOTA dataset, with a total of 140 epochs. For the HRSC2016 and UCAS-AOD datasets, the batch sizes are set to 32, and the initial learning rates are set to $2\times 10^{-4}$ and $1\times 10^{-4}$, respectively.

\noindent\textbf{Model Evaluation Metrics.} Following the conventions in the field of object detection, we adopt the standard COCO format mean Average Precision (mAP) as the evaluation metric.
\subsection{Object detection in aerial images}
For the experiments on DOTA, we adopt data augmentation with random rotations at angles of $\{0, \pi/2, \pi, 3\pi/2\}$ and category balancing, along with horizontal flipping. Since aerial images typically vary significantly in size, this leads to increased computational and memory requirements. To address this issue, the preprocessing of remote sensing images employs the DOTA\_devkit toolbox, using a detection strategy that involves tiling the images into slices to alleviate this challenge, even if the slices contain no detectable objects.
\begin{figure}[!t]
	\captionsetup{labelsep=period}
	\raggedleft 
	\begin{minipage}[t]{1\textwidth} 
		\centering
		\includegraphics[width=\linewidth]{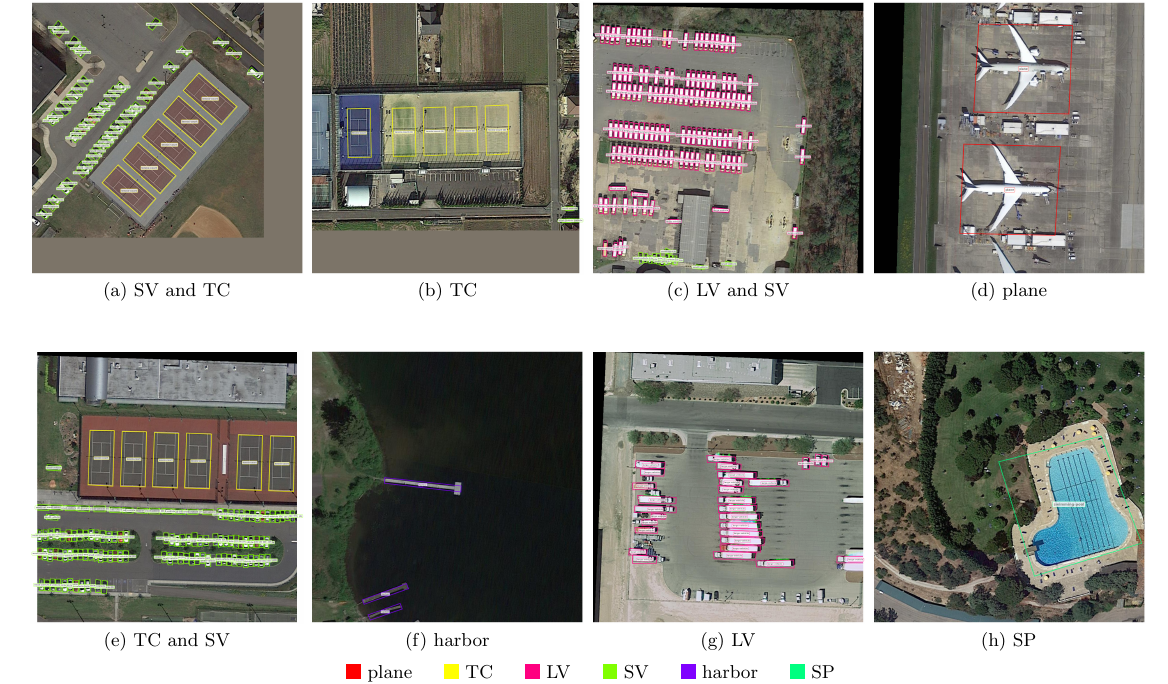}
		\caption{Visualization of detection results on the DOTA dataset, demonstrating the model's performance on large-scale, obliquely oriented, and densely arranged objects. The examples include various object categories such as planes, large vehicles (LV), small vehicles (SV), tennis courts (TC), harbors, and swimming pools (SP).}
		\label{fig:2x4_grid}
	\end{minipage}
	\vspace{-10pt} 
\end{figure}
We first investigated different settings for the image input size and the overlap stride during cropping. As shown in Table 2, we report the mAP during inference. Initially, we cropped the images into $1024 \times 1024$ slices. When the overlap stride was increased from 200 to 512, the mAP improved from 70.38\% to 72.19\%. This improvement is attributed to the fact that many large objects and those located near image boundaries become incomplete due to cropping, making it difficult to detect objects near the edges under the original cropping strategy, which directly impacts detection accuracy. However, increasing the overlap stride from 200 to 512 also increased the number of slice images from 8,143 to 20,012, leading to an overall increase in inference time by approximately 175\%. This is because the greater overlap between slices results in more slices, inevitably extending inference time. When we employed half-precision floating-point (FP16) to accelerate inference, we were able to reduce the inference time by 256 seconds, with an mAP of 71.45\%. The experimental results demonstrate that our method is both efficient and effective, and our detection strategy achieves a better trade-off between detection speed and detection accuracy.

\noindent\textbf{Visualization of Overall Results.} Some qualitative results on DOTA and HRSC2016 are shown in Figure 4, respectively. We display all detected targets whose classification scores exceed a certain threshold. From the detection results, it can be observed that our model is effective in detecting large-sized objects and densely arranged objects.

\subsection{Qualitative Results}
\label{Qualitative Results}
Figure 4 presents representative visual results on the DOTA dataset. As illustrated, compared to the previous Rotation RetinaNet method without the MSK Module, our proposed RMK RetinaNet demonstrates strong adaptability to scene variations, effectively handling significant size differences among target objects. It ensures reliable detection of larger objects (e.g.,ST) while maintaining focus on smaller ones (e.g., Ship,LV,SV) and avoiding erroneous detection outcomes(e.g.,SV).
\begin{figure*}[!t]
	\captionsetup{labelsep=period}
	\centering
	\includegraphics[width=1.0\linewidth]{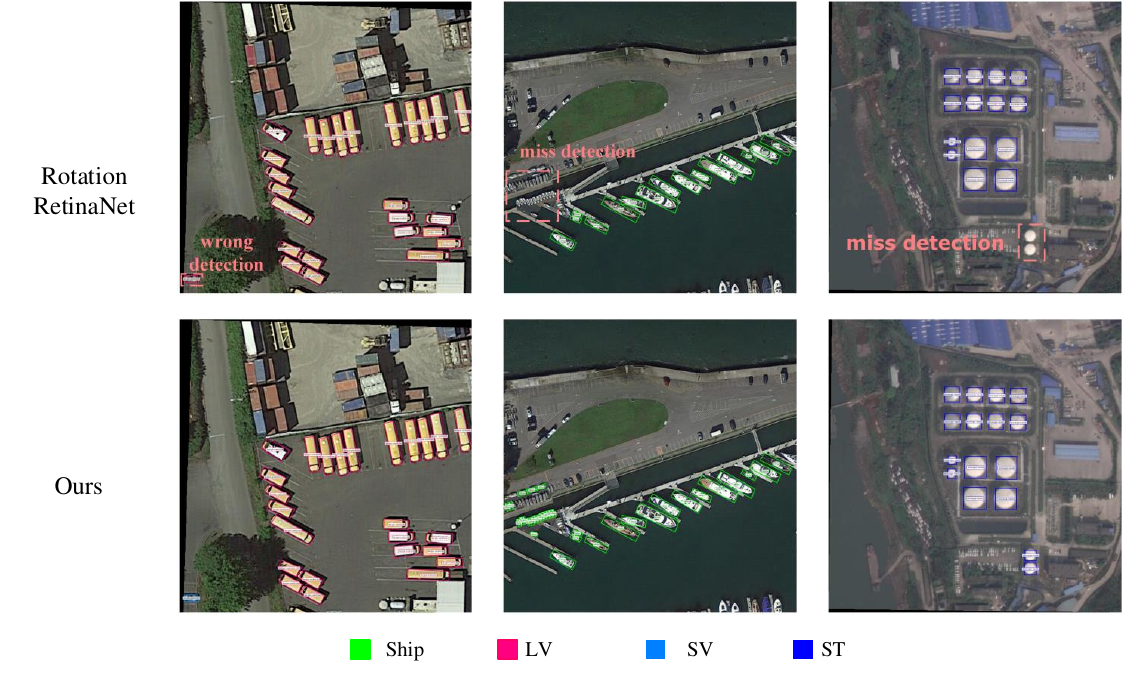}
	\caption{Qualitative comparison of detection results on the DOTA dataset. Top row: Rotation RetinaNet exhibits wrong detections and missed detections; Bottom row: our RMK RetinaNet successfully detects all instances, including ships, large vehicles (LV), small vehicles (SV), and storage tanks (ST). See \textcolor{red}{\S\ref{Qualitative Results}} for details.}
	\label{fig:msk_architecture}
\end{figure*}

\subsection{Ablation Experiment}
In this section, we conducted a series of experiments on the DOTA test set to validate the effectiveness of our method. Across all experiments, ResNet50 was employed as the backbone network. Rotation RetinaNet was chosen as the baseline in this paper. As a single-stage detector, Rotation RetinaNet is sufficiently efficient. As shown in Table 2, Rotation RetinaNet achieves 68.49\% mAP, demonstrating the reliability of our baseline. To verify the effectiveness of each module modification in RMK RetinaNet, training was performed on the DOTAv1.0 dataset at a resolution of 1024 × 1024 pixels. Each variant of the solution underwent a dedicated training scheme for 140 epochs, designed to balance the pursuit of high accuracy with resource efficiency.

\noindent\textbf{Effectiveness of MSK Module.} 
As described in Section 3.2, the MSK Module was added and compared with the baseline model to validate its effectiveness. While keeping all other settings unchanged, we solely incorporated the MSK Module into the baseline model. As shown in Table 2, the MSK Module achieves an approximately 0.41\% improvement in mAP compared to the baseline model.This demonstrates that in remote sensing object detection, multi-scale convolutional kernels with an Inception-like structure are not merely a “trick” for boosting performance but represent an effective solution addressing a core challenge in this field. By expanding the receptive field, this approach equips the model with the ability to accurately represent and recognize various targets in complex scenes.
\begin{table}[!t]
	\centering
	\small
	\setlength{\tabcolsep}{0.34pt}
	 \captionsetup{labelsep=period}
	\caption{Ablation study of key components in Rotated Multi-Kernel RetinaNet. We evaluate the contribution of each module by progressively adding MSK Module, MDCAA Module, Bottom-up Path Module, and Euler Angle Encoding Module. The results show that the full combination achieves the best mAP of 70.38\%.}
	\label{tab:results}
	\input{./Tables1/table2.tex}

\end{table}

\noindent\textbf{Effectiveness of the MDCAA Module.}\ As shown in Table 2, by introducing both the MSK Module and the MDCAA Module to the baseline model, our method achieves a significant performance improvement, with a mAP gain of approximately 0.71\%. This result validates that, firstly, the MSK Module can extract features accurately; secondly, the MDCAA Module can effectively enhance the orientation sensitivity of features. The combination of these two modules contributes to the notable increase in detection accuracy.

\noindent\textbf{Effectiveness of the Bottom-up Path Module.} As described in Section 3.4, we introduced the bottom-up path module and compared it with the baseline model to validate its effectiveness. With all other settings unchanged, we added only the bottom-up path module to the baseline model, which resulted in an approximately 0.3\% improvement in mAP. The experimental results demonstrate that the bottom-up path module is effective for fusing shallow positional information with deep semantic information.

\noindent\textbf{Effectiveness of Euler Angle Encoding Module.}
According to our experimental results, it was found that the standalone Euler Angle Encoding Module improves upon the baseline model by 0.46\%. This is sufficient to demonstrate the effectiveness of the module in angle encoding and decoding.

Furthermore, we further integrated the MSK Module, MDCAA Module, and Bottom-up Path Module into the baseline model, with the experimental results presented in Table 2. The experiments show that our model achieves a 1.1\% improvement in detection performance compared to the baseline model. Ultimately, our integrated model incorporating all modules achieves a 1.89\% improvement over the baseline model.This is because the MSK Module emphasizes its refined capability in extracting key information, while the MDCAA Module enhances orientation awareness while filtering out irrelevant or disruptive noise features. Furthermore, the Bottom-up Path ensures the flow of low-level information to higher layers, transmitting positional information and guaranteeing effective fusion between low-level positional details and high-level semantic information.
\subsection{Comparison with Advanced Technologies}
In this section, the proposed RMK RetinaNet is compared with other state-of-the-art methods on the  HRSC2016, DOTA, and UCAS-AOD datasets. The experimental settings for these comparisons have been described in Section 4.1.
\begin{figure}[!t]
	\captionsetup{labelsep=period}
	\centering
	\begin{minipage}[t]{0.8\textwidth}
		\centering
		\includegraphics[width=\linewidth]{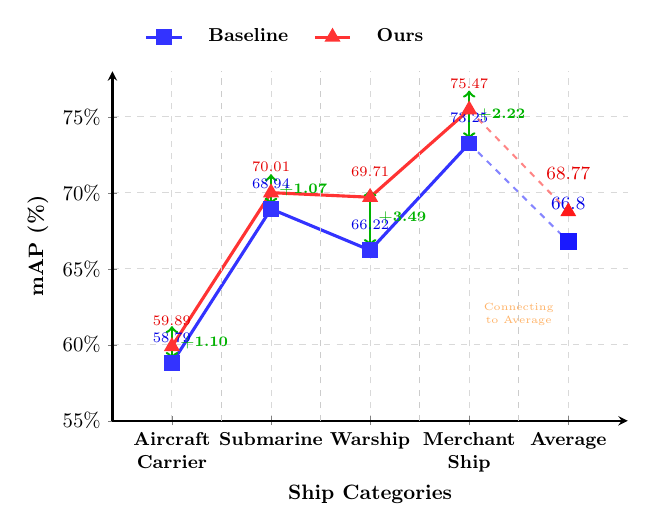}
		\caption{Performance comparison of object detection on the HRSC2016 dataset. We divide the HRSC2016 dataset into 4 categories, and our method achieves 68.77\% mAP, significantly outperforming the baseline model.}
		\label{fig:Detection Performance on HRSC2016}
	\end{minipage}
	\vspace{-10pt}
\end{figure}

\begin{table*}[!t]
	\captionsetup{labelsep=period}
	\centering
	\caption{Detection accuracy comparison on DOTA dataset. For each method, we report Average Precision (AP) for 15 object categories: Plane, Baseball Diamond (BD), Bridge, Ground Track Field (GTF), Small Vehicle (SV), Large Vehicle (LV), Ship, Tennis Court (TC), Basketball Court (BC), Storage Tank (ST), Soccer Ball Field (SBF), Roundabout (RA), Harbor, Swimming Pool (SP), and Helicopter (HC). The overall mean AP (mAP) is also provided. Our method achieves the best performance with 70.38\% mAP, demonstrating significant improvements over previous approaches.}
	\label{tab:dota_results}
	\input{./Tables1/table3.tex}
\end{table*}

\begin{table}[t]
	\captionsetup{labelsep=period}
	\centering
	\small
	\setlength{\tabcolsep}{0.34pt}
	\caption{Performance evaluation on the UCAS-AOD dataset. We compare our method against YOLOv2, R-DFPN, DRBox, and Rotation RetinaNet. Results are reported in terms of mAP and per-class AP for Plane and Car. Our approach achieves the best overall performance with 91.735\% mAP.}
	\label{tab:ucas_results}
	\input{./Tables1/table4.tex}
\end{table}
\noindent\textbf{Results on HRSC2016.} As shown in Figure 5,, although our Rotated Multi-Kernel RetinaNet outperforms existing methods on this dataset, the detection performance appears less satisfactory due to the fact that the dataset does not consolidate 4 subcategories into a single "ship" category for training and evaluation, as done by the vast majority of other methods. However, if we adopt the same approach as other methods by merging the 4 subcategories into a single "ship" category for both training and evaluation, our chosen baseline model achieves an accuracy of 82.51\%. In comparison, our proposed RMK RetinaNet shows an improvement of 1.52\% in accuracy over the baseline model.

\noindent\textbf{Results on DOTA.} The experimental results are compared with the comprehensive performance of various detectors in Table 3. It should be noted that due to differences in image resolution, network architecture, detection framework, training strategies, and optimization techniques employed across various methods, their performance often varies. This diversity in technical approaches makes it somewhat challenging to conduct an absolutely fair comparison among different methods. Despite the aforementioned challenges, the proposed method achieves competitive detection performance compared to the various detection methods in Table 3, without employing any test-time data augmentation or multi-scale training and testing strategies. On the DOTA-v1.0 dataset, our method attains an mAP of 70.38\%.

\noindent\textbf{Results on UCAS-AOD.} According to the experimental data in Table 4, which evaluates the performance of various published methods on the UCAS-AOD dataset, we found that the accuracy for detecting objects like aircraft is significantly higher than that for detecting vehicles. Our results are the best among all published methods, achieving a mAP of 91.73\%.
Building upon existing research on the precise representation of rotated bounding boxes in remote sensing object detection, this study innovatively incorporates prior knowledge specific to remote sensing scenarios and proposes a new remote sensing object detection algorithm called RMK RetinaNet (Rotated Multi‑Kernel RetinaNet). Using the single-stage Rotation RetinaNet as the baseline, this algorithm enhances multi-scale feature fusion through a MSK Module, designs a feature pyramid based on a MDCAA mechanism to address challenges such as arbitrary orientations, varying scales, and complex backgrounds of objects in remote sensing images, and introduces a Bottom-up Path Module to strengthen the expression of low-level localization information at higher layers. Additionally, an EAEM is incorporated to further optimize the directional representation of rotated object detection boxes.
\section{Conclusion}
Experiments on mainstream remote sensing object detection datasets, including DOTA, HRSC2016, and UCAS‑AOD, demonstrate that RMK RetinaNet shows no significant performance gap compared to multiple state-of-the-art detection algorithms, validating the effectiveness of incorporating remote sensing prior knowledge and the practical value of the proposed modules. Future work will further explore the generalization capability of the model in larger-scale and more diverse remote sensing scenarios and investigate lightweight designs to meet practical deployment requirements.




\bibliographystyle{elsarticle-num}  
\bibliography{main}         
\end{document}

%% file: Tables1/table1.tex
  \captionsetup{labelsep=period}
	\centering
	\caption{Computational cost analysis of our method with PKI and MSK modules on DOTA. FLOPs are measured with 1024$\times$1024 input resolution.}
	\begin{tabular}{c|cccc}
		\toprule
		\textbf{Model} & \textbf{Input Size} & \textbf{FLOPs} & \textbf{Params}\\
		\midrule
		Ours+PKI Module &1024$\times$1024 & 65.693 & 33.612\\
		Ours+MSK Module &1024$\times$1024 & 65.693 & 33.558\\
		\bottomrule
	\end{tabular}
	\label{tab:model_comparison}

%% file: Tables1/table2.tex
	\begin{minipage}{\textwidth}
		\centering
		\begin{tabular}{c|*{6}{>{\centering\arraybackslash}p{1.2cm}}}  
			\toprule
			\multicolumn{7}{c}{\textbf{Different Settings of Rotated Multi-Kernel RetinaNet}} \\ 
			\midrule
			MSK Module          & $\checkmark$ & $\times$ & $\times$ & $\checkmark$ & $\checkmark$ & \textbf{$\checkmark$} \\
			MDCAA Module           & $\times$     & $\times$ & $\times$     & $\checkmark$ & $\checkmark$ & \textbf{$\checkmark$} \\
			Bottom-up path Module & $\times$   & $\checkmark$ & $\times$   & $\times$     & $\checkmark$ & \textbf{$\checkmark$ }\\ 
			Euler Angle Encoding Module & $\times$    & $\times$ & $\checkmark$ & $\times$     & $\times$     & \textbf{$\checkmark$} \\ 
			\midrule
			mAP                 & 68.90 & 68.79 & 68.95 & 69.20 & 69.59 & \textbf{70.38} \\ 
			\bottomrule
		\end{tabular}%
		\label{tab:settings}
	\end{minipage}

%% file: Tables1/table3.tex
	\resizebox{\textwidth}{!}{%
		\setlength{\tabcolsep}{3pt}
		\renewcommand{\arraystretch}{0.9}
		\begin{tabular}{|c|c|c|c|c|c|c|c|c|c|c|c|c|c|c|c|c|c|c|}
			\hline
			\footnotesize Methods & \footnotesize Backbone & \footnotesize Plane & \footnotesize BD & \footnotesize Bridge & \footnotesize GTF & \footnotesize SV & \footnotesize LV & \footnotesize Ship & \footnotesize TC & \footnotesize BC & \footnotesize ST & \footnotesize SBF & \footnotesize RA & \footnotesize Harbor & \footnotesize SP & \footnotesize HC & \footnotesize mAP\\
			\hline
			\footnotesize R-FCN\cite{dai2016rfcn} & \footnotesize ResNet-101 & \footnotesize 81.01 & \footnotesize 58.96 & \footnotesize 31.64 & \footnotesize 58.97 & \footnotesize 49.77 & \footnotesize 45.04 & \footnotesize 49.29 & \footnotesize 68.99 & \footnotesize 52.07 & \footnotesize 67.42 & \footnotesize 41.83 & \footnotesize 51.44 & \footnotesize 45.15 & \footnotesize 53.30 & \footnotesize 33.89 & \footnotesize 52.58\\
			\footnotesize IENet\cite{lin2019ienet} & \footnotesize ResNet-101 & \footnotesize 80.20 & \footnotesize 62.54 & \footnotesize 39.82 & \footnotesize 32.07 & \footnotesize 49.71 & \footnotesize 65.01 & \footnotesize 52.58 & \footnotesize 81.45 & \footnotesize 44.66 & \footnotesize 78.51 & \footnotesize 46.54 & \footnotesize 56.73 & \footnotesize 64.40 & \footnotesize 64.24 & \footnotesize 36.75 & \footnotesize 57.14\\
			\footnotesize R-DFPN\cite{yang2018automatic} & \footnotesize ResNet-101 & \footnotesize 80.92 & \footnotesize 65.82 & \footnotesize 33.77 & \footnotesize 58.94 & \footnotesize 55.77 & \footnotesize 50.94 & \footnotesize 54.78 & \footnotesize 90.33 & \footnotesize 66.34 & \footnotesize 68.66 & \footnotesize 48.73 & \footnotesize 51.76 & \footnotesize 55.10 & \footnotesize 51.32 & \footnotesize 35.88 & \footnotesize 57.94\\
			\footnotesize PIoU\cite{chen2020piou} & \footnotesize DLA-34 & \footnotesize 80.9 & \footnotesize 69.70  & \footnotesize 24.10 & \footnotesize 60.20 & \footnotesize 38.30 & \footnotesize 64.40 & \footnotesize 64.80 & \footnotesize 90.90 & \footnotesize 77.20 & \footnotesize 70.40 & \footnotesize 46.50 & \footnotesize 37.10 & \footnotesize 57.10 & \footnotesize 61.90 & \footnotesize 64.00& \footnotesize 60.50\\
			\footnotesize R$^{2}$CNN\cite{jiang2017r2cnn} & \footnotesize ResNet-101 & \footnotesize 80.94 & \footnotesize 65.67 & \footnotesize 35.34 & \footnotesize 67.44 & \footnotesize 59.92 & \footnotesize 50.91 & \footnotesize 55.81 & \footnotesize 90.67 & \footnotesize 66.92 & \footnotesize 72.39 & \footnotesize 55.06 & \footnotesize 52.23 & \footnotesize 55.14 & \footnotesize 53.35 & \footnotesize 48.22 & \footnotesize 60.67\\
			\footnotesize RRPN\cite{ma2018arbitrary} & \footnotesize ResNet-101 & \footnotesize 88.52 & \footnotesize 71.20 & \footnotesize 31.66 & \footnotesize 59.30 & \footnotesize 51.85 & \footnotesize 56.19 & \footnotesize 57.25 & \footnotesize 90.81 & \footnotesize 72.84 & \footnotesize 67.38 & \footnotesize 56.69 & \footnotesize 52.84 & \footnotesize 53.08 & \footnotesize 51.94 & \footnotesize 53.58 & \footnotesize 61.01\\
			\footnotesize Deformable R-FCN\cite{dai2017deformable} & \footnotesize ResNet-101 &\footnotesize 87.97 & \footnotesize 76.69 & \footnotesize 46.97 & \footnotesize 68.76 & \footnotesize 55.86 & \footnotesize 63.5 & \footnotesize 56.9 & \footnotesize 90.13 & \footnotesize 75.81 & \footnotesize 64.8 & \footnotesize 51.73 & \footnotesize 60.01 & \footnotesize 74.96 & \footnotesize 71.41 & \footnotesize 52.95 & \footnotesize 66.56 \\
			\footnotesize CADNet\cite{zhang2019cadnet} & \footnotesize ResNet-101 & \footnotesize 87.80 & \footnotesize 82.40 & \footnotesize 49.40 & \footnotesize 73.50 & \footnotesize 71.10 & \footnotesize 63.50 & \footnotesize 76.60 & \footnotesize 90.90 & \footnotesize 79.20 & \footnotesize 73.30 & \footnotesize 48.40 & \footnotesize 60.90 & \footnotesize 62.00 & \footnotesize 67.00 & \footnotesize 62.20 & \footnotesize 69.90\\
			RoI Trans.\cite{ding2019learning} & \footnotesize ResNet-101 & \footnotesize 88.64 & \footnotesize 78.52 & \footnotesize 43.44 & \footnotesize 75.92 & \footnotesize 68.81 & \footnotesize 73.68 & \footnotesize 90.74 & \footnotesize 77.27 & \footnotesize 81.46 & \footnotesize 86.15 & \footnotesize 58.39 & \footnotesize 53.54 & \footnotesize 62.83 & \footnotesize 58.93 & \footnotesize 47.67 & \footnotesize 69.56\\
			\footnotesize Azimi et al.\cite{azimi2018towards}& \footnotesize ResNet-101 & \footnotesize 81.36 & \footnotesize 74.30 & \footnotesize 47.70 & \footnotesize 70.32 & \footnotesize 64.89 & \footnotesize 67.82& \footnotesize 69.98 & \footnotesize 90.76 & \footnotesize 79.06 & \footnotesize 78.20 & \footnotesize 53.64 & \footnotesize 62.90& \footnotesize 67.02 & \footnotesize 64.17 & \footnotesize 50.23 & \footnotesize 68.16\\
			\footnotesize Rotation RetinaNet\cite{zan2019automatic} & \footnotesize ResNet-50& \footnotesize 88.01 & \footnotesize 78.01 & \footnotesize 43.81 & \footnotesize 76.12 & \footnotesize 68.00 & \footnotesize 73.05 & \footnotesize 72.91 & \footnotesize 89.74 & \footnotesize 76.61 & \footnotesize 81.25 & \footnotesize 58.19 & \footnotesize 53.04 & \footnotesize 62.38 & \footnotesize 58.64 & \footnotesize 47.6 & \footnotesize 68.49\\
			\hline
			\footnotesize \textbf{ours} & \footnotesize ResNet-50& \footnotesize \textbf{89.90} & \footnotesize \textbf{79.90} & \footnotesize \textbf{45.70} & \footnotesize \textbf{78.01} & \footnotesize \textbf{69.89} & \footnotesize \textbf{74.91} & \footnotesize \textbf{74.83} & \footnotesize \textbf{91.63} & \footnotesize \textbf{78.50} & \footnotesize \textbf{83.14} & \footnotesize \textbf{60.08}& \footnotesize \textbf{54.93} & \footnotesize \textbf{64.27} & \footnotesize \textbf{60.53} & \footnotesize \textbf{49.49} & \footnotesize \textbf{70.38}\\
			\hline
		\end{tabular}%
	}
	

%% file: Tables1/table4.tex
\label{tab:detection_performance}
\setlength{\tabcolsep}{0.8em}    
\begin{tabular}{p{5cm}ccc}
	\toprule
	\textbf{Method} & \textbf{mAP} & \textbf{Plane} & \textbf{Car} \\
	\midrule
	YOLOv2 \cite{redmon2017yolo9000}     & 87.90        & 96.60          & 79.20        \\
	R-DFPN \cite{yang2018automatic}     & 89.20        & 95.90          & 82.50        \\
	DRBox \cite{liu2017learning}      & 89.95        & 94.90          & 85.00        \\
	Rotation RetinaNet\cite{zan2019automatic}    &   91.495        & 96.12        &   86.87\\
	\textbf{Ours}     &   \textbf{91.735}     & \textbf{97.01} & \textbf{86.46} \\
	\bottomrule
\end{tabular}